\documentclass[10pt,twocolumn,letterpaper]{article}

\usepackage{wacv}
\usepackage{times}
\usepackage{epsfig}
\usepackage{graphicx}
\usepackage{amsmath}
\usepackage{amssymb}
\usepackage{algorithm}
\usepackage{algpseudocode}

\def\wacvPaperID{710} 

\wacvfinalcopy 

\ifwacvfinal
\def\assignedStartPage{9876} 
\fi

\ifwacvfinal
\usepackage[breaklinks=true,bookmarks=false]{hyperref}
\else
\usepackage[pagebackref=true,breaklinks=true,colorlinks,bookmarks=false]{hyperref}
\fi

\ifwacvfinal
\else
\fi

\begin{document}

\title{Autoencoding Video Latents for Adversarial Video Generation}

\author{Sai Hemanth Kasaraneni\\
Samsung Research Institute\\
Noida, India\\
{\tt\small k.saihemant@samsung.com}
}

\maketitle

\begin{abstract}
Given the three dimensional complexity of a video signal, training a robust and diverse GAN based video generative model is onerous due to large stochasticity involved in data space. Learning disentangled representations of the data help to improve robustness and provide control in the sampling process. For video generation, there is a recent progress in this area by considering motion and appearance as orthogonal information and designing architectures that efficiently disentangle them. These approaches rely on handcrafting architectures that impose structural priors on the generator to decompose appearance and motion codes in the latent space. Inspired from the recent advancements in the autoencoder based image generation, we present AVLAE (\textbf{A}dversarial \textbf{V}ideo \textbf{L}atent \textbf{A}uto\textbf{E}ncoder) which is a two stream latent autoencoder where the video distribution is learned by adversarial training. In particular, we propose to autoencode the motion and appearance latent vectors of the video generator in the adversarial setting. We demonstrate that our approach learns to disentangle motion and appearance codes even without the explicit structural composition in the generator. Several experiments with qualitative and quantitative results demonstrate the effectiveness of our method.
\end{abstract}

\section{Introduction}

Generative Adversarial Networks (GAN)~\cite{GAN} are implicit generative models which have shown remarkable progress in image generation. Many variants of this framework were proposed to address training instability~\cite{ImprovedGAN,WGAN,WGAN-GP,GANReg,SNGAN} and to improve the image generation quality~\cite{DCGAN,LAPGAN,PGAN,SAGAN,BigGAN,StyleGAN,StyleGAN2}. More applications of GANs include image translation~\cite{pix2pix,CycleGAN} and super-resolution~\cite{SRGAN,StarGAN}. Videos are three dimensional spatio-temporal signals with different objects performing diverse actions over time. However video generation is more challenging than image generation due to the additional temporal dimension. The generated videos should not only maintain consistent geometrical structure, composition / appearance of various objects in the video across all frames but also the motion of different objects over time should be physically plausible. The motion of the several objects can be highly stochastic~\cite{SVGLP} or governed by complex dynamics between frames~\cite{UCF}. Increase in model complexity and memory~\cite{TGAN2} are the other consequences. Recent GAN based approaches decomposed the generation pipeline into output foreground - background~\cite{VGAN} or the latent static appearance-motion~\cite{MoCoGAN,G3AN} to tackle with these challenges. The former way poses a static background assumption while the latter approach requires carefully designed architectures to learn disentangled representations\footnote{\label{note1}G$^3$AN~\cite{G3AN} carries a three-stream generator with ~25M trainable parameters to generate $64\times64$ resolution videos with 16 frames}. In this work, we aim to achieve motion and appearance disentanglement without imposing strict structural priors on the generator. We propose to learn the disentangled representations from the encoded discriminative features of the generated videos.

Autoencoders on the other hand, are encoder-decoder type networks which learn representations to the data in unsupervised manner. Variational AutoEncoders (VAE)~\cite{VAE} are the generative autoencoders where the encoder approximates the posterior over latent variables, which is being utilized for sampling latent codes to be processed by the decoder to output the reconstructions. Another class of generative models~\cite{AAE,ABP,BiGAN,ALI,AGE,IntroVAE,ALAE} aim to combine adversarial training with autoencoders to mitigate with 1. mode-collapse in conventional GAN 2. workaround for reconstruction loss in pixel space and 3. to give generative capability to the normal autoencoders. Pidhorskyi \etal~\cite{ALAE} proposed a latent autoencoder for image generation that can leverage advantages of GANs and provide less entangled representations of the images. They applied non-linear transformation of latent vectors sampled from fixed prior to an intermediate latent space free from the fixed distribution imposition, with only restriction being to match encoded features used in the discriminator. Such unconstrained latent space is shown to learn rich semantic representations~\cite{StyleGAN,StyleGAN2} as its distribution is allowed to be learnt from the data itself. In this work, we propose a generative video autoencoder (AVLAE), extension of~\cite{ALAE} for video generation. We reconstruct the two stream intermediate latent space by computing the appearance and motion (optical flow) cues from the generated videos.  Our main contributions are as follows:

\begin{enumerate}
    \item We propose a novel generative video latent autoencoder framework for unconditional video generation
    \item We propose a simple and straight-forward technique for efficient motion and appearance disentanglement with a rudimentary generator architecture
    \item We conduct extensive evaluation of our method along with ablation studies and compare to the state-of-the-art video generation methods
\end{enumerate}

\section{Related work}

\subsection{Video Generation}

Our problem mainly concerns with unconditional video generation and learning disentangled representations. Existing unconditional video generation approaches are mainly based on GANs. GAN based video generative models~\cite{VGAN,TGAN,MoCoGAN,FTGAN,PSWG,TGAN2,DVDGAN,LDVD-GAN,G3AN,GIG} employ adversarial training for learning video distribution. Vondrick \etal~\cite{VGAN} train a two stream (3D spatio-temporal foreground and 2D spatial background streams) generator to output the final video. But this comes with an assumption that the background remains static through-out the video which might not be true for complex datasets~\cite{UCF}. Saito \etal~\cite{TGAN} implemented a temporal generator producing sequence of latent vectors and an image generator transforming each latent vector to an image. Tulyakov \etal~\cite{MoCoGAN} generated video clips by traversing through the decomposed latent space of image generator. Each latent vector is a concatenation of content vector (constant through out the video) and motion vector (sampled from motion subspace at each timestep). They also introduced the convention to use two discriminators (video and image discriminators) playing the minimax game concurrently with the generator. Ohnishi \etal~\cite{FTGAN} generated videos in two stages by first generating optical flow from noise and further adding texture to synthesize intact video. TGAN2~\cite{TGAN2} was optimized on memory and computational cost by instituting temporal sub-sampling layers at multiple levels of generation process. G$^3$AN~\cite{G3AN} decomposes motion and appearance in a three stream generator with a main spatio-temporal stream and two auxillary (spatial and temporal) streams and outperformed the previous works. Despite learning disentangled representations, G$^3$AN needed careful construction of a three stream generator and a factorized spatio-temporal attention module. We rather consider a rudimentary spatio-temporal generator mapping two independent and identically distributed latent vectors to output space.


\subsection{Conditional Generation}

As video generation from random noise vectors require to model the dataset distribution, many methods were proposed for generating videos by leveraging conditional inputs with recurrent architectures. Video prediction~\cite{DeepMS,MCNet,longterm,DRNet,SV2P,SVGLP,SAVP,DDPAE,Yu2020Efficient,Jin_2020_CVPR} task particularly learns a generative model by conditioning on starting $K>0$ frames of the target video. Other modes of input include conditioning on semantic segmentation map~\cite{segmentmap}, start and end frames~\cite{p2p} and captions~\cite{captions}. However the presence of additional inputs makes it different from the unconditional video generation problem where the inputs are random vectors drawn from latent space.

\subsection{Adversarial Generative Autoencoders}

There have been recent efforts in combining adversarial training with traditional autoencoders~\cite{AE} for image generation. Makhzani \etal~\cite{AAE} performed variational inference by imposing an arbitrary prior distribution using adversarial training over aggregated posterior and prior latent distributions. Larsen \etal~\cite{ABP} used adversarial learning in output space of a VAE and replaced pixel reconstruction loss with feature-wise errors in the discriminator. Donahue \etal~\cite{BiGAN} and Dumoulin \etal~\cite{ALI} concurrently trained an auxiliary network that maps generated samples to the latent features of the generator. The discriminator then distinguishes the joint distribution of latent variables and image samples from generator and auxiliary networks. Pidhorskyi \etal~\cite{ALAE} proposed Adversarial Latent AutoEncoder (ALAE) which generated high fidelity images that are comparable to GANs~\cite{PGAN,StyleGAN,StyleGAN2} while learning less entangled representations. However as per our knowledge, this is the first attempt to train a adversarial autoencoder that efficiently learns 3 dimensional video distribution while also decomposing the motion and appearance features in latent space.

\begin{figure*}
\begin{center}
\includegraphics[width=0.9\linewidth]{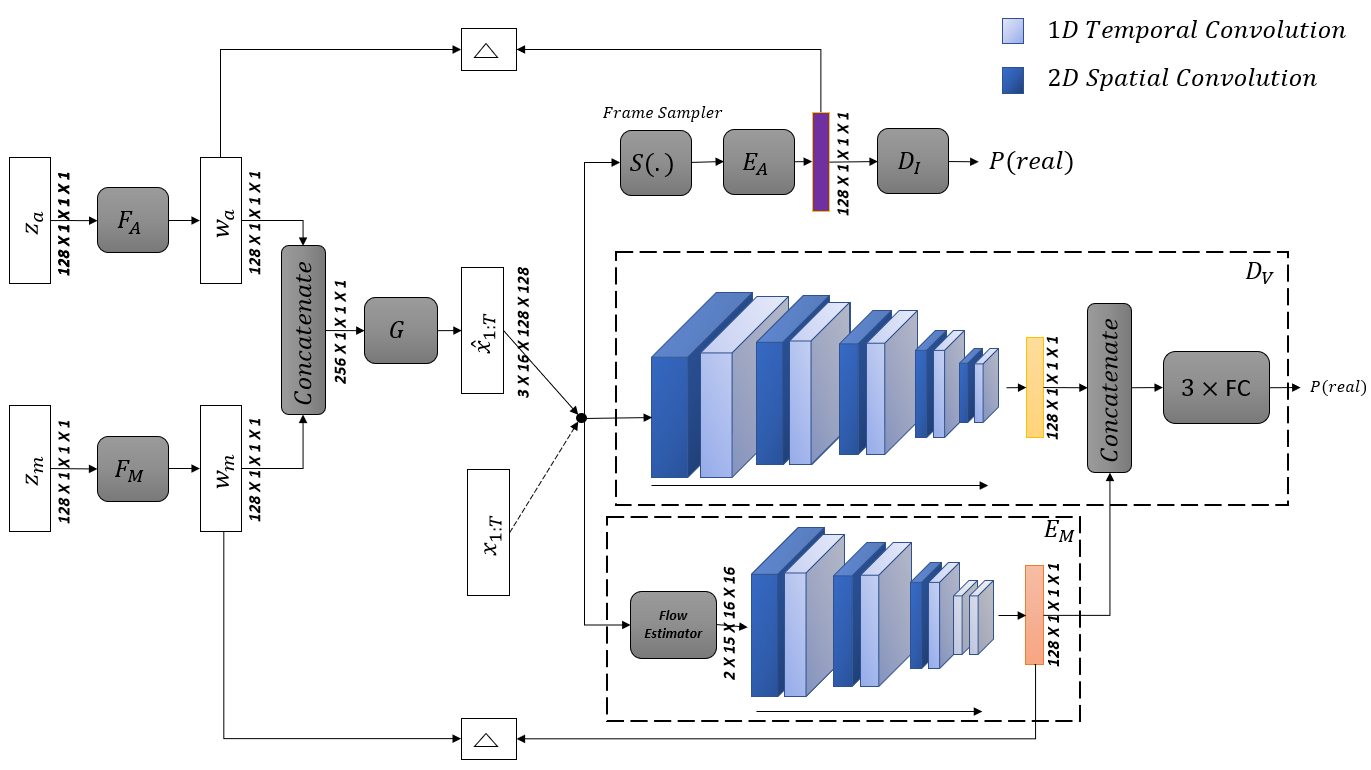}
\end{center}
   \caption{\textbf{AVLAE} architecture (best viewed in color). $F_A$ and $F_M$ project input appearance and motion latent vectors onto intermediate latent space. The video generator $G$ outputs generated video of resolution $128 \times 128 \times 16$ from the concatenated intermediate latent vector. The motion encoder $E_M$ and appearance encoder $E_A$ encodes motion, appearance characteristics to latent space respectively. The output dimensions are of the form $channels \times depth \times Height \times Width$. $\Delta$ denotes L1 reconstruction loss.}
\label{fig1}
\end{figure*}

\section{GAN Preliminaries}

A GAN contains two networks, a generator $G$ and a discriminator $D$, which are trained in adversarial fashion. The generator is trained to model the data distribution and generate realistic samples by taking noise vector $z$ sampled from fixed prior $P_{Z}$ (usually gaussian), as input. Concurrently discriminator which takes the generated samples $G(z)$ and samples from data distribution $P_{\mathcal{D}}$, is trained to distinguish between them and also provides feedback to the generator.

\begin{equation}
\begin{split}
    \min_{G} \max_{D} \mathcal{L}(G,D) = \mathbb{E}_{x \sim P_{\mathcal{D}}(x)}[\log D(x)] +\\
    \mathbb{E}_{z \sim P_Z(z)}[\log (1-D(G(z)))]\label{eq1}
\end{split}
\end{equation}

The training of a GAN includes optimizing eq.\ref{eq1} which resembles a zero-sum game between generator and discriminator until they converge to a stable nash equilibrium. With enough capacity of $G$ and $D$, it was shown that the generator distribution $P_{G}$ converges to the real data distribution $P_{\mathcal{D}}$ allowing us to sample novel and realistic data samples~\cite{GAN}.

\section{Approach}

In this section, we introduce the architecture of the proposed model. We decompose the generator of a typical GAN into three networks $F_A$, $F_M$ and $G$. $F_A$ and $F_M$ are the non-linear transformation functions for appearance and motion latent vectors respectively. They map the respective latent vectors sampled from input latent subspace to the intermediate latent subspace. $G$ is a spatio-temporal generator which outputs the generated video of fixed number of frames. On the other side, we retain the video discriminator $D_V$~\cite{MoCoGAN,G3AN} and decompose the image discriminator into the appearance encoder $E_A$ and the image latent discriminator $D_I$. The image discriminator requires generator to output videos with coherent spatial structures in each frame while video discriminator requires the generated videos to be spatio-temporally consistent. We also introduce the motion encoder $E_M$ network which explicitly encodes the dynamics of the input video to a bottleneck representation. see Fig.\ref{fig1}

\subsection{Video Generator}

We implement $F_A$ and $F_M$ as stacked fully connected layers with leaky relu activations between them and $G$ as a simple spatio-temporal upsampling generator with factorized (1+2)D transpose convolutions~\cite{STConv}. $F_A$ projects the input vector $z_A$ sampled from normal distribution onto the intermediate appearance subspace $W_A$. Similarly $F_M$ projects another independent vector $z_M$ sampled from identical distribution onto the intermediate motion subspace $W_M$. We aim $W_A$ to learn motion independent appearance features of the video and $W_M$ to learn disentangled motion characteristics. Appearance and motion projections are then concatenated to form samples in intermediate video latent space $W_V$. i.e. $W_V$ = $W_A \times W_M$. Network $G$ takes the concatenated vector and translates to a video of fixed frame length, $T$. So the generated video distribution would be

\begin{equation}
\begin{split}
    P_{G}(x) = \int_{w_{A}}\int_{w_{M}}P_{G}(x/w_A,w_M)\,P_{F_{A}}(w_A)\\
    P_{F_{M}}(w_M) \,dw_{A}\,dw_{M} \label{eq2}
\end{split}
\end{equation}

\subsection{Image and Motion Encoders}

Given an input video, we want our motion encoder $E_{M}$ to explicitly encode the dynamics of the video in a representational space. Optical Flow is the motion of image pixel intensities between the frames in a scene which can be attributed to motion of objects in the scene. It carries low-level motion information which can further be utilized in high-level vision tasks like video recognition~\cite{VAR} and even in video generation~\cite{FTGAN}. Optical flow estimation with deep neural networks has achieved great progress over the traditional methods~\cite{flownet,flownet2,RAFT}. Our motion encoder first estimates the optical flow sequence of length $T-1$ for the input video of $T$ frames with a pre-trained flow prediction network, which is further processed by factorized spatio-temporal convolution layers to output the encoded motion vector. Thus obtained encoded motion vector is concatenated before the fully connected layers of video discriminator as shown in Fig.\ref{fig1} to ensure temporal coherence in generated videos. We used the state-of-the-art end-to-end differentiable flow estimator RAFT~\cite{RAFT} pretrained on sintel dataset~\cite{sintel} for obtaining optical flow at $1/8^{th}$ of the spatial resolution of the input video. The distribution at the output of motion encoder for input generated video is

\begin{equation}
\begin{split}
    P_{E_{M}}(w^{\prime}_{M}) = \int_{x}P_{E_{M}}(w^{\prime}_{M}/x)\,P_{G}(x)\,dx \label{eq3}
\end{split}
\end{equation}

We also sample randomly a single frame from the input video and feed it to the appearance encoder $E_A$ to get appearance latent vector. The encoded image is further passed to $D_I$ to get real/fake probabilities. If the distribution of all generated frames is $P_{G}(x_{t})$,

\begin{equation}
\begin{split}
    P_{E_{A}}(w^{\prime}_{A}) = \int_{x_{t}}P_{E_{A}}(w^{\prime}_{A}/x_{t})\,P_{G}(x_{t})\,dx_{t} \label{eq4}
\end{split}
\end{equation}

would be the output distribution of $E_{A}$. We observe that by training with adversarial objective, the generator distribution converges towards the real data distribution. i.e. $P_{G}(x) \approx P_{\mathcal{D}}(x)$ and $P_{G}(x_t) \approx P_{\mathcal{D}}(x_t)$, where $P_{\mathcal{D}}(x)$ and $P_{\mathcal{D}}(x_t)$ are real video and real frame distribution respectively. So by replacing $P_{\mathcal{D}}(x)$, $P_{\mathcal{D}}(x_t)$ for $P_{G}(x)$, $P_{G}(x_t)$ in eq.~\ref{eq3},\ref{eq4} respectively, the encoder distributions converge to same for real video/frame and generated video/frame inputs.

Now the intermediate latent distributions $P_{F_A}(w_A)$, $P_{F_M}(w_M)$, $P_{E_A}(w^{\prime}_A)$ and $P_{E_M}(w^{\prime}_M)$ are not enforced to follow any predefined prior and are free to learn from the data. Additionally we put the constraint that the latent space pairs $(P_{F_A}(w_A), P_{F_M}(w_M))$ and $(P_{E_A}(w^{\prime}_A), P_{E_M}(w^{\prime}_M))$ should be same. That is,

\begin{equation}
\begin{split}
    P_{F_A}(w_A) = P_{E_A}(w^{\prime}_A) \label{eq5}\\
    P_{F_M}(w_M) = P_{E_M}(w^{\prime}_M)
\end{split}
\end{equation}

With this constraint, the networks $G$ - $(E_A,E_M)$ become \textit{generator - two stream encoder} type autoencoder which autoencodes the joint video latent space $W_V$. We implement this constraint by a reconstruction loss between $(w_A,w_M)$ and $(w^{\prime}_A,w^{\prime}_M)$ which is minimized by updating $G$,$E_M^*$ and $E_A$ where $E_M^*$ represents motion encoder excluding the flow estimator network.

\begin{algorithm}[b!]
\caption{AVLAE Training}
\label{alg1}
\footnotesize
\begin{algorithmic}[1]
\State \(\theta_{E_M^*} \gets \theta_{E_M} - params(Flow Estimator)\) 
\State \(\theta_{F_A}, \theta_{F_M}, \theta_G, \theta_{E_A}, \theta_{E_M^*}, \theta_{D_I}, \theta_{D_V} \gets\) Initialize network parameters
\While{not converged}
\State Step I. Update $E_A$, $E_M^*$, $D_I$ and $D_V$ 
\State \(x \gets\) Random mini-batch from dataset \(\mathcal{D}\)
\State \(z_A \gets\) Samples from prior \(P_{Z_A} = \mathcal{N}(0,I)\)
\State \(z_M \gets\) Samples from prior \(P_{Z_M} = \mathcal{N}(0,I)\)
\State \(\hat{x} \gets G(F_A(z_A),F_M(z_M))\)
\State \(x_t \sim \{x_{1:T}\}\) and \( \hat{x}_t \sim \{\hat{x}_{1:T}\} \)
\State \(L^{E_M^*,D_V}_{adv} \gets \log(D_V(x,E_M(x))) + \log(1 - D_V(\hat{x},E_M(\hat{x}))) \)
\State \(L^{E_A,D_I}_{adv} \gets \log(D_I(E_A(x_t))) + \log(1 - D_I(E_A(\hat{x_t}))) \)
\State \(grad \gets \nabla_{\theta_{E_M^*}, \theta_{D_V}}L^{E_M^*,D_V}_{adv}\)
\State \( \theta_{E_M^*}, \theta_{D_V} \gets \textsc{Adam}(grad, \theta_{E_M^*}, \theta_{D_V}, \alpha, \beta_1, \beta_2)\)
\State \(grad \gets \nabla_{\theta_{E_A}, \theta_{D_I}} L^{E_A,D_I}_{adv}\)
\State \( \theta_{E_A}, \theta_{D_I} \gets \textsc{Adam}(grad, \theta_{E_A}, \theta_{D_I}, \alpha, \beta_1, \beta_2)\)

\State Step II. Update $F_A$, $F_M$ and $G$ 
\State \(z_A \gets\) Samples from prior \(P_{Z_A} = \mathcal{N}(0,I)\)
\State \(z_M \gets\) Samples from prior \(P_{Z_M} = \mathcal{N}(0,I)\)
\State \(\hat{x} \gets G(F_A(z_A),F_M(z_M)) \)
\State \( \hat{x}_t \sim \{\hat{x}_{1:T}\} \)
\State \(L^{F_A,F_M,G}_{adv} \gets \log(1 - D_I(E_A(\hat{x}_t))) + \log(1 -D_V(E_M(\hat{x}))) \)
\State \(grad \gets \nabla_{\theta_{F_A}, \theta_{F_M}, \theta_G} L^{F_A,F_M,G}_{adv}\)
\State \( \theta_{F_A}, \theta_{F_M}, \theta_G \gets \textsc{Adam}(grad, \theta_{F_A}, \theta_{F_A}, \theta_G, \alpha, \beta_1, \beta_2)\)

\State Step III. Update $E_A$, $E_M^*$ and $G$
\State \(z_A \gets\) Samples from prior \(P_{Z_A} = \mathcal{N}(0,I)\)
\State \(z_M \gets\) Samples from prior \(P_{Z_M} = \mathcal{N}(0,I)\)
\State \(L^{E_A,E_M^*,G}_{rec} \gets \| F_M(z_M) - E_M(G(F_A(z_A),F_M(z_M))) \|_2^2 + \| F_A(z_A) - E_A(G(F_A(z_A),F_M(z_M))_{t \sim \{1:T\}}) \|_2^2\)
\State \(grad \gets \nabla_{\theta_{E_A}, \theta_{E_M^*}, \theta_G} L^{E_A,E_M^*,G}_{rec}\)
\State \(\theta_{E_A}, \theta_{E_M^*}, \theta_G \gets \textsc{Adam}(grad, \theta_{E_A}, \theta_{E_M^*}, \theta_G, \alpha, \beta_1, \beta_2)\)
\EndWhile
\end{algorithmic}
\end{algorithm}

\subsection{Learning}

Let $T > 0$ be the number of frames to be generated and $x_{1:T}$ represents the video sample of T frames from the dataset $\mathcal{D}$. Let $z_{A}$ and $z_{M}$ are appearance and motion latent vector samples from input latent subspaces $Z_A$ and $Z_M$ respectively. Then the adversarial objective of our framework is

\begin{equation}
\begin{split}
    \min_{F_A,F_M,G} \,\,\max_{E_A,E_M^*,D_I,D_V} \mathcal{L}_{adv}(F_A,F_M,G,E_A,E_M,D_I,D_V)\label{eq6}
\end{split}
\end{equation}

where 
\begin{equation}
\begin{split}
    \mathcal{L}_{adv}(F_A,F_M,G,E_A,E_M,D_I,D_V) =\\ \mathcal{L}_I(F_A,F_M,G,E_A,D_I) + \mathcal{L}_V(F_A,F_M,G,E_M,D_V)\label{eq7}
\end{split}
\end{equation}

$\mathcal{L}_I$ and $\mathcal{L}_V$ are the adversarial losses from $D_I$ and $D_V$ respectively. Let $\hat{x}_{1:T} = G(F_A(z_A),F_M(z_M))$ be the generated video.

\begin{equation}
\begin{split}
     \mathcal{L}_V(F_A,F_M,G,E_M,D_V) = \mathbb{E}_{x}[\log D_V(x,E_M(x))] +\\
    \mathbb{E}_{z_A,z_M}[\log (1-D_V(\hat{x},E_M(\hat{x})))]\label{eq8}
\end{split}
\end{equation}

\begin{equation}
\begin{split}
    \mathcal{L}_I(F_A,F_M,G,E_A,D_I) = \mathbb{E}_{x_t}[\log D_I(E_A(x_t))] +\\
    \mathbb{E}_{z_A,z_M}[\log (1-D_I(E_A(\hat{x}_{t \sim \{1:T\}})))]\label{eq9}
\end{split}
\end{equation}

where $\mathbb{E}_x$, $\mathbb{E}_{z_A,z_M}$, $\mathbb{E}_{x_t}$ are short-hands for $\mathbb{E}_{x \sim P_{\mathcal{D}}(x)}$, $\mathbb{E}_{z_A \sim P_{Z_A}(z_A),z_M \sim P_{Z_M}(z_M)}$ and $\mathbb{E}_{x_t \sim P_{\mathcal{D}}(x_t)}$ respectively. We also additionally optimize over the following objective to enforce the constraint eq.\ref{eq5}.

\begin{equation}
\begin{split}
    \min_{E_A,E_M^*,G}\, \mathcal{L}_{rec}(G,E_A,E_M) = K_1\| F_M(z_M) - E_M(\hat{x}) \|_2^2 + \\
    K_2\| F_A(z_A) - E_A(\hat{x}_{t \sim \{1:T\}}) \|_2^2\label{eq10}
\end{split}
\end{equation}

We update the respective networks to optimize w.r.t. eq.\ref{eq6},\ref{eq10} in alternative steps in each iteration. We summarized our training algorithm in alg.\ref{alg1}. We note that the flow estimator is not updated in any of the steps and only passes the gradients to the generator.

\section{Experiments}

We compare our approach with state-of-the art GAN based video generation methods \textbf{VGAN}~\cite{VGAN}, \textbf{TGAN}~\cite{TGAN}, \textbf{MoCoGAN}~\cite{MoCoGAN} and \textbf{G$^3$AN}~\cite{G3AN} which are closest to our work, qualitatively and quantitatively. Then we conduct experiments to demonstrate the motion and content disentanglement by manipulating the latent spaces. Finally we perform ablation studies for our proposed method to show the effectiveness of each component of the architecture.

\subsection{Datasets}

We evaluate the proposed method on three datasets namely \textbf{UvA-NEMO} smiles~\cite{UvA}, \textbf{Weizmann Human Actions}~\cite{WA} and \textbf{UCF-101} Action Recognition~\cite{UCF} datasets.

UvA-NEMO smiles dataset consists of 1240 spontaneous/posed smiling video sequences spanning around 400 identities. The Weizmann Human Actions dataset comprises 90 videos of 9 people performing 10 natural actions. Similar to~\cite{G3AN}, we augment the dataset by horizontally flipping the videos. The UCF-101 is a more diverse 101 category actions dataset with videos collected from YouTube. We resize each frame to $170\times128$ resolution and center crop to $128\times128$.

\subsection{Implementation Details}
\label{sec:ID}

We trained our model using PyTorch for generating $128\times128$ spatial resolution videos with $T = 16$ frames. We used ADAM optimizer~\cite{Adam} with $\beta_1 = 0.5$ and $\beta_2 = 0.999$ and a learning rate of $2e^{-4}$ for all networks. All our latent vectors $z_A$,$z_M$,$w_A$ and $w_M$ are 128 dimensional vectors. We used $K_1 = K_2 = 1$ in eq.\ref{eq10} in all our experiments. The networks $F_A$,$F_M$ are independent and implemented as 5 stacked fully connected layers each while $D_I$ and $D_V$ are 3 stacked fully connected layers each. Through out our framework, we used factorized versions of 3D convolutional and 3D transpose convolutional layers.

\subsection{Quantitative Analysis}

We evaluate each method quantitatively with two metrics which are video extensions of widely used image generation evaluation metrics, Fr\'echet Inception Distance (FID)~\cite{FID} on all aforementioned datasets and Inception Score (IS)~\cite{ImprovedGAN} on UCF-101 dataset. We utilize the features of 3D CNN feature extractor~\cite{FE} similar to~\cite{G3AN}, on dataset and generated video samples. The FID-Score compares for samples from dataset and generated videos, the first and second moments of the distribution of 3D CNN features. The Inception score (IS) is calculated as relative entropy between conditional and marginal distributions of classification labels for generated video samples. The FID score captures the video quality and temporal coherence. Higher inception score implies low entropy in conditional and high entropy in marginal class distributions. So IS captures both visual quality and diversity of generated videos.

We use our trained model to generate videos randomly which are to be used in FID and IS calculation. As previous methods were trained for $64\times64$ resolution videos, we also downsample our generated videos to $64\times64$ and report FID values on both resolutions to avoid unwarranted advantanges simply because of higher resolution. We randomly generate 5000 video samples for each trained model and utilize them in calculation of the metrics. FID and IS scores for different approaches are shown in Table.\ref{tab1} and Table.\ref{tab2} correspondingly. \textsc{G$^3$AN} results are different from those reported in~\cite{G3AN} as the authors of \textsc{G$^3$AN} have improved their model in their official code\footnote{\label{note2}\url{https://github.com/wyhsirius/g3an-project}}.

\begin{table}
\begin{center}
\begin{tabular}{l|c|c|c}
Method & \textbf{UvA} & \textbf{Weizmann} & \textbf{UCF-101} \\
\hline
VGAN & 235.01 & 158.04 & 115.06 \\
TGAN & 216.41 & 99.85 & 110.58\\
MoCoGAN & 197.32 & 92.18 & 104.14\\
G$^3$AN & 64.82 & 76.94 & 87.20 \\
\textbf{AVLAE-64} & 60.05 & 68.46 & 84.38\\
\textbf{AVLAE} & \textbf{54.20} & \textbf{65.71} & \textbf{80.59}\\
\end{tabular}
\end{center}
\caption{\textbf{Video FID score} comparison with \textbf{state-of-the-art} approaches (lower is better). All results except ours are taken from~\cite{G3AN}.}
\label{tab1}
\end{table}

\begin{table}
\begin{center}
\begin{tabular}{c|c|c}
VGAN & TGAN & MoCoGAN \\
2.94 & 2.74 & 3.06 \\
\hline
G$^3$AN & \textbf{AVLAE-64} & \textbf{AVLAE} \\
3.88 & 3.93 & \textbf{3.97} \\
\end{tabular}
\end{center}
\caption{\textbf{Video IS score} comparison with \textbf{state-of-the-art} approaches (higher is better) on \textbf{UCF-101} dataset. All results except ours are taken from~\cite{G3AN}.}
\label{tab2}
\end{table}

\begin{figure*}
\begin{center}
\includegraphics[width=0.9\linewidth]{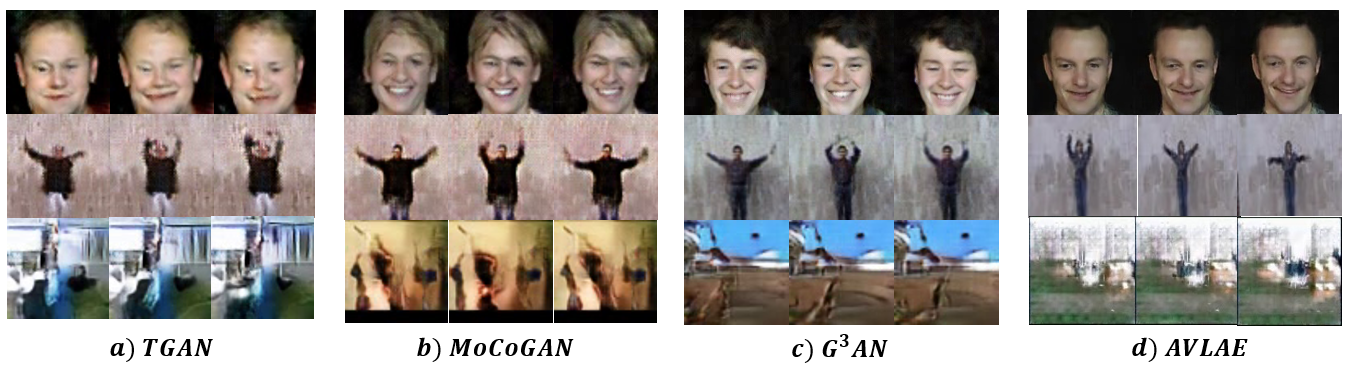}
\end{center}
   \caption{\textbf{Visual comparison} of generated samples ($1^{st}$,$10^{th}$ and $16^{th}$ frames) with \textbf{TGAN}, \textbf{MoCoGAN} and \textbf{G$^3$AN} methods. Top row are generated samples from \textbf{UvA-NEMO}, middle row are \textbf{Weizmann Action} and \textbf{UCF-101} samples in third row. Video frames of TGAN, G$^3$AN and MoCoGAN are from~\cite{G3AN}.}
\label{fig2}
\end{figure*}

Our approach consistently outperformed the previous methods in all datasets with respect to FID score. This manifests the superior visual quality and spatio-temporal consistency of our generated video samples. The higher inception score on UCF-101 demonstrates the diverse video samples spanning the latent space. We note that G$^3$AN produced second best results and is closest to our approach. However, the G$^3$AN packs an expensive generator with around 25M trainable parameters for generating $64\times64\times16$ videos while our generator contains nearly same number of trainable parameters for generating $128\times128\times16$ resolution videos.

\subsection{Qualitative Analysis}

We randomly sample the videos from the trained models by sampling the latent vectors and feeding them to the generator. We present generated samples by different methods on the three datasets in Fig.\ref{fig2}.

\begin{figure}[t]
\begin{center}
\includegraphics[width=0.9\linewidth]{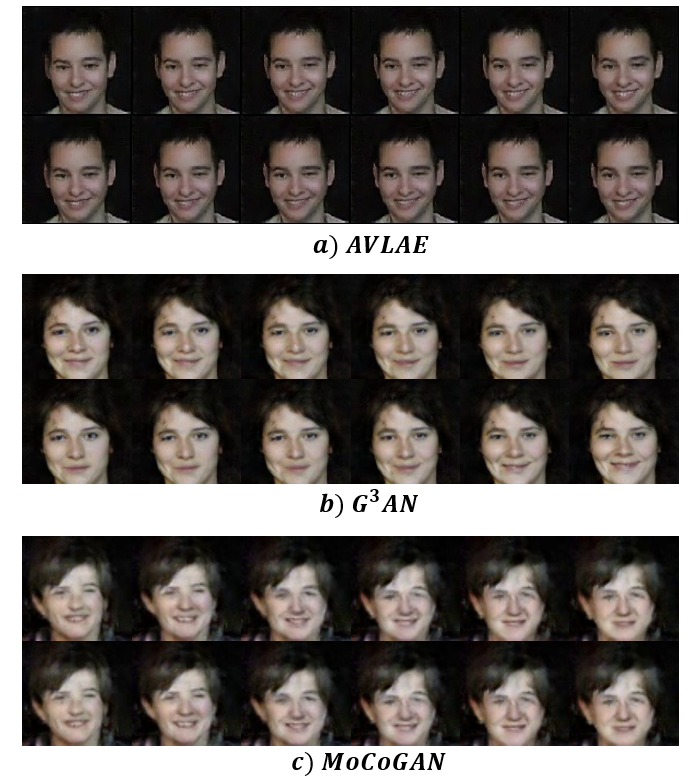}
\end{center}
   \caption{Video samples with \textbf{same appearance code} $z_A$ and two \textbf{different samples of $z_M$} for each method on UvA-NEMO dataset. Best viewed when zoomed-in}
\label{fig3}
\end{figure}

\begin{figure}[t]
\begin{center}
\includegraphics[width=0.9\linewidth]{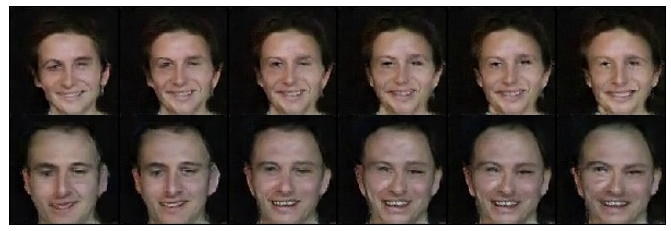}
\end{center}
   \caption{Video samples generated by AVLAE with \textbf{same motion code $z_M$} and two \textbf{different samples of $z_A$}. Best viewed when zoomed-in}
\label{fig4}
\end{figure}

We experiment with the motion and appearance disentanglement in the latent space. Particularly, we fix one of the input latent vectors $z_A,z_M$ while varying the other and visualize the videos obtained in the process. Fig.\ref{fig3} shows the samples obtained while changing the motion code $z_M$ and keeping the appearance code $z_A$ same. All the three methods including ours, preserved the appearance characteristics (facial appearance) across the two samples but the motion in the samples generated by MoCoGAN is unaltered even though $z_M$ is changed. G$^3$AN and AVLAE are able to alter the motion (amount of smile in this case) for different samples of $z_M$. This shows the efficient motion disentanglement in latent space. We note that unlike G$^3$AN, there is no explicit structural composition of generator to achieve the same in our framework (both $z_M$ and $z_A$ are equivalent with respect to the generator's architecture). Figure~\ref{fig4} further shows the alteration of appearance characteristics in generated videos by our model while changing $z_A$ without altering $z_M$.

\begin{figure}[t]
\begin{center}
\includegraphics[width=0.9\linewidth]{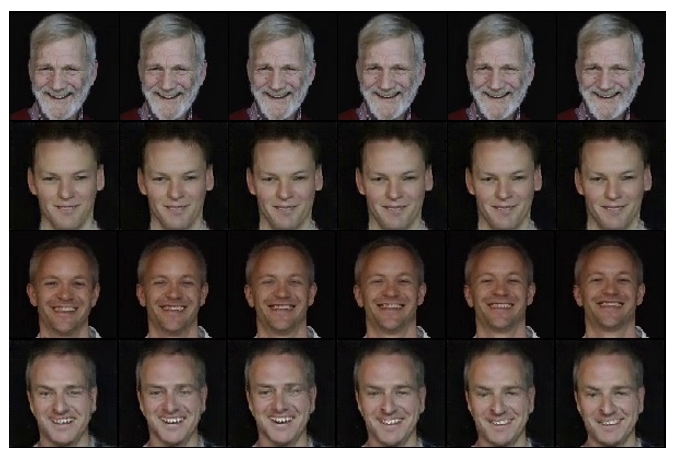}
\end{center}
   \caption{\textbf{Reconstructions} of real video samples. $1^{st}$ and $3^{rd}$ rows are the real video inputs. $2^{nd}$ and $4^{th}$ row videos are their corresponding reconstructions.}
\label{fig5}
\end{figure}

With the input video sample from dataset distribution, the reconstructions given by the autoencoder can also be visualized by using the feature representations. For input real video $x$, we sample a single frame $x_{t \sim \{1:T\}}$ and construct the model output as $G(E_A(x_t),E_M(x))$. Fig.~\ref{fig5} shows such reconstructions for held out videos during training.

\subsection{Ablation Study}

Further we study the contribution of each individual component in our proposed architecture. As our main contributions are regarding latent reconstruction and motion encoder block, we retrain our model on UvA-NEMO dataset in the following four cases of modification : i) removing latent motion code $w_M$ reconstruction ii) removing appearance latent $w_A$ reconstruction iii) removing both appearance and motion latent reconstruction iv) removing both appearance and motion latent reconstructions and Motion Encoder $E_M$. The first three cases are equivalent to choosing hyperparameters $K_1=0$ , $K_2=0$ and $K_1 = K_2 = 0$ respectively. Note that these three cases still include the motion encoder in their architecture. In fourth case, we remove motion encoder additionally and change the number of channels of fully connected layers in $D_V$ accordingly which results in normal video GAN with image and video discriminators. We evaluate these cases quantitatively in terms of FID scores on UvA-NEMO dataset in Table. \ref{tab3}.

\begin{table}
\begin{center}
\begin{tabular}{l|c|c|c}
Architecture & \textbf{FID} \\
\hline
AVLAE & 54.20\\
$-w_A\ rec$ & \textbf{37.50}\\
$-w_M\ rec$ & 59.84\\
$-w_A\ rec$, $-w_M\ rec$  & 59.28\\
$-w_A\ rec$, $-w_M\ rec$, $-E_M$ & 233.82\\
\end{tabular}
\end{center}
\caption{\textbf{Video FID score} comparison on UvA-NEMO dataset for different modifications in architecture}
\label{tab3}
\end{table}

\begin{figure}[t]
\begin{center}
\includegraphics[width=0.9\linewidth]{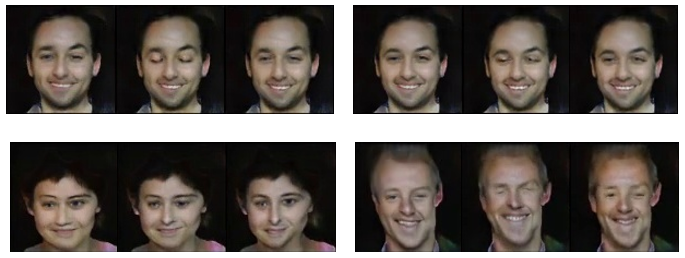}
\end{center}
   \caption{\textbf{Ablation Study}: First row shows two videos with different $z_A$ and same $z_M$ when $w_A$ reconstruction is removed. Second row shows two videos with same $z_A$ and different $z_M$ when $w_M$ reconstruction is removed.}
\label{fig6}
\end{figure}

We first observe that the motion encoder has the most effect on generating high fidelity videos with temporal consistency, as removing the motion encoder resulted in the highest rise in FID score. Further, the removal of appearance reconstruction gave best FID score overall. This is in-line with the findings of~\cite{ALAE} that the intermediate latent space learning more discriminative features from image discriminator might not be 'rich' enough. However the removal of appearance reconstruction resulted in the loss of control over the latent space as in, the appearance characteristics remain unaltered even when the $z_A$ is altered as shown in Fig.~\ref{fig6}. Similarly we observe the appearance alteration while altering the motion vector $z_M$ when $w_M$ reconstruction is removed. This might be because of the feedback from video discriminator to the motion encoder forcing the latent subspace $W_M$ to learn both motion and appearance features when either of the latent spaces are left unconstrained. However removing the latent motion reconstruction also resulted in videos with unnatural motion between the frames. The intact AVLAE architecture generates high quality videos while also decomposing the motion and appearance.

\section{Conclusion}

In this paper, we have presented an adversarial autoencoder framework for video generation. Our work is the first autoencoder based video generative model that is able to learn video distribution efficiently and perform better than existing GAN based video generation methods. We considered a rudimentary structure for the generator and proposed to autoencode the decomposed latent vectors. The experiments demonstrated the ability of our framework to generate high fidelity videos with spatio-temporal consistency as well as to learn less entangled representations of the videos in the latent space resulting in decomposition of motion and appearance. Further ablation experiments prove that the intact AVLAE model gives more control over the latent space while preserving the video quality.

{\small
\bibliographystyle{ieee_fullname}
\bibliography{main}
}

\end{document}